%% file: anonymous-submission-latex-2025.tex
\documentclass[letterpaper]{article} % DO NOT CHANGE THIS
\usepackage{aaai25}
\usepackage{times}  % DO NOT CHANGE THIS
\usepackage{helvet}  % DO NOT CHANGE THIS
\usepackage{courier}  % DO NOT CHANGE THIS
\usepackage[hyphens]{url}  % DO NOT CHANGE THIS
\usepackage{graphicx} % DO NOT CHANGE THIS
\usepackage{natbib}  % DO NOT CHANGE THIS AND DO NOT ADD ANY OPTIONS TO IT
\usepackage{caption} % DO NOT CHANGE THIS AND DO NOT ADD ANY OPTIONS TO IT
\usepackage{algorithm}
\usepackage{algorithmic}
\usepackage{amsmath}
\usepackage[caption=false,font=normalsize,labelfont=sf,textfont=sf]{subfig}
\usepackage{textcomp}
\usepackage{url}
\usepackage{verbatim}
\usepackage{cite}
\usepackage[usenames,dvipsnames,svgnames,table]{xcolor}
\usepackage{xspace}
\usepackage{amsthm,amsmath,amssymb,lipsum}
\usepackage{mathrsfs}
\usepackage{booktabs}
\usepackage{multirow}
\usepackage{multicol}
\usepackage{bbding}
\usepackage{amsmath}
\usepackage{array}
\usepackage[all,cmtip]{xy}

\usepackage{mathtools}
\usepackage{pdfpages}

\usepackage{newfloat}
\usepackage{listings}
\DeclareCaptionStyle{ruled}{labelfont=normalfont,labelsep=colon,strut=off} % DO NOT CHANGE THIS
\floatstyle{ruled}
\newfloat{listing}{tb}{lst}{}
\floatname{listing}{Listing}
\def\onedot{\ifx\@let@token.\else.\null\fi\xspace}
\def\eg{\emph{e.g}\onedot} 
\def\eg{\emph{e.g}\onedot} 
\def\ie{\emph{i.e}\onedot}

\def\etal{\emph{et al}\onedot}
\makeatother

\def\Vec#1{{\boldsymbol{#1}}}

\usepackage{amsthm}

\title{PEARL: Input-Agnostic Prompt Enhancement with Negative Feedback Regulation for Class-Incremental Learning}
\author{
    Yongchun Qin\textsuperscript{\rm 1,\rm 2}, Pengfei Fang\textsuperscript{\rm 1,\rm 2}$^*$, Hui Xue\textsuperscript{\rm 1,\rm 2}\thanks{Corresponding author.}\\
}
\affiliations{
    \textsuperscript{\rm 1}School of Computer Science and Engineering, Southeast University, Nanjing 210096, China\\
    \textsuperscript{\rm 2}Key Laboratory of New Generation Artifcial Intelligence Technology and Its Interdisciplinary\\
    Applications (Southeast University), Ministry of Education, China\\
    \{ycqin, fangpengfei, hxue\}@seu.edu.cn
}

\usepackage{bibentry}
\begin{document}

\maketitle

\begin{abstract}

% Class-incremental learning (CIL) aims to continuously introduce novel categories into a classification system without forgetting previously learned ones, thus adapting to evolving data distributions. Researchers are now focusing on leveraging the rich semantic information of pre-trained models (PTMs) in CIL tasks. Prompt learning has been adopted in CIL for its ability to adjust data distribution to better align with pre-trained knowledge. This paper critically examines, from the perspective of prompt learning, the limitations of existing methods that heavily rely on input information.
% To address this issue, we propose a novel method for PTM-based CIL called input-agnostic \underline{\textbf{{P}}}rompt \underline{\textbf{{E}}}nhancement with neg\underline{\textbf{{A}}}tive feedback \underline{\textbf{{R}}}egu\underline{\textbf{{L}}}ation (\textbf{PEARL}). In PEARL, we implement an input-agnostic global prompt coupled with an adaptive momentum update strategy to reduce the model's dependency on data distribution, thereby effectively mitigating catastrophic forgetting. This adaptive momentum update, guided by negative feedback regulation, addresses the parameter sensitivity inherent in fixed-weight momentum updates. Furthermore, it fosters the continuous enhancement of the prompt to new tasks by harnessing correlations between different tasks in CIL.	Experiments on six benchmarks demonstrate that our method achieves state-of-the-art performance. The code will be released.	

Class-incremental learning (CIL) aims to continuously introduce novel categories into a classification system without forgetting previously learned ones, thus adapting to evolving data distributions. Researchers are currently focusing on leveraging the rich semantic information of pre-trained models (PTMs) in CIL tasks. Prompt learning has been adopted in CIL for its ability to adjust data distribution to better align with pre-trained knowledge. This paper critically examines the limitations of existing methods from the perspective of prompt learning, which heavily rely on input information. To address this issue, we propose a novel PTM-based CIL method called Input-Agnostic \underline{\textbf{{P}}}rompt \underline{\textbf{{E}}}nhancement with Neg\underline{\textbf{{A}}}tive Feedback \underline{\textbf{{R}}}egu\underline{\textbf{{L}}}ation (\textbf{PEARL}). In PEARL, we implement an input-agnostic global prompt coupled with an adaptive momentum update strategy to reduce the model's dependency on data distribution, thereby effectively mitigating catastrophic forgetting. Guided by negative feedback regulation, this adaptive momentum update addresses the parameter sensitivity inherent in fixed-weight momentum updates. Furthermore, it fosters the continuous enhancement of the prompt for new tasks by harnessing correlations between different tasks in CIL. Experiments on six benchmarks demonstrate that our method achieves state-of-the-art performance. The code is available at: https://github.com/qinyongchun/PEARL.

% The advent of pre-trained models (PTMs) has significantly impacted the realm of computer vision.	

% To address this issue, we propose a momentum updating strategy inspired by negative feedback regulation. This strategy reframes knowledge retention from a mere constraint in CIL to a critical prerequisite.
% It ensures that the model acquires new knowledge only after effectively retaining existing knowledge, thus allowing continuous enhancement of prompts.

\end{abstract}

%%%%%%%%%%%%%%%%%%%%%%%%%%%%%%%%%%%%%%%%%%%%%%%%%%%%%%%%%%%%%%%%%%%%%%%%%%%%%%%%%%%%%%%%%

\input{m_Intro}

\input{m_Related}

\input{m_method}
\input{m_Expt}
\input{m_Conclusion}

% \clearpage
%%%%%%%%%%%%%%%%%%%%%%%%%%%%%%%%%%%%%%%%%%%%%%%%%%%%%%%%%%%%%%%%%%%%%%%%%%%%%%%%%%%%%%%%%

\bibliography{aaai25}

\newpage


\section*{Supplementary material}

# Supplementary Material for
# PEARL: Input-Agnostic Prompt Enhancement with Negative Feedback Regulation for Class-Incremental Learning

Yongchun Qin[1,2], Pengfei Fang[1,2], Hui Xue[1,2*]

[1]School of Computer Science and Engineering, Southeast University, Nanjing 210096, China
[2]Key Laboratory of New Generation Artifcial Intelligence Technology and Its Interdisciplinary
Applications (Southeast University), Ministry of Education, China
{ycqin, fangpengfei, hxue}@seu.edu.cn

## Abstract

In the main paper, we propose the Input-Agnostic **P**rompt **E**nhancement with Neg**A**tive Feedback **R**egu**L**ation (**PEARL**) for PTM-based CIL. In this supplementary material, we provide additional details, organized as follows:

- Section 1 provides the pseudo code and corresponding equations to enhance readability.
- Section 2 provides a detailed comparison between continual positional encoding and the proposed segmented positional encoding.
- Section 3 provides more details on integrating global prompts into existing "query-select" based methods.
- Section 5 reports additional ablation study, mainly focusing on hyperparameters.

## 1 Pseudo Code for PEARL

We summarize the specific pipeline of PEARL in Algorithm 1 to more clearly illustrate its update process. To enhance readability, we have listed the relevant equations below.

PEARL can be divided into two phases based on the session during incremental learning. When $t=1$, PEARL does not involve momentum updating, only the prompt token needs to be calculated:

$$[\text{PT}]_1^t, [\text{SP}]_1^t = \mathcal{E}_1\big(\text{ConCat}([\text{PT}]_0^t, \mathcal{P}[1:\frac{\mathcal{M}}{\mathcal{N}}\times t])\big),$$
$$[\text{PT}]_2^t, [\text{SP}]_2^t = \mathcal{E}_2\big(\text{ConCat}([\text{PT}]_1^t, [\text{SP}]_1^t)\big),$$
$$\vdots$$
$$[\text{PT}]_i^t, [\text{SP}]_i^t = \mathcal{E}_i\big(\text{ConCat}([\text{PT}]_{i-1}^t, [\text{SP}]_{i-1}^t)\big). \quad (1)$$

Based on the prompt token, for any instance $\boldsymbol{x}$, logits can be calculated as follows:

$$\boldsymbol{l}^1 = g \circ \mathcal{V}\big(\boldsymbol{x}, \{[\text{PT}]_i^1\}\big). \quad (2)$$

PEARL updates the prompt pool and prompt encoder based on Cross-Entropy loss:

$$\mathcal{L}_{\text{cls}} = \mathbb{E}_{(\boldsymbol{x},y)\sim\mathbf{D}^t}\text{CE}(\boldsymbol{l}^t, y). \quad (3)$$

In the second phase (*i.e.* $t > 1$), PEARL needs to update both the prompt token and $\alpha^\tau$, respectively. The momentum prompt is obtained as follows:

$$[\text{PT}]_i^{\text{mem}} = \alpha^\tau \cdot [\text{PT}]_i^{t-1} + (1-\alpha^\tau) \cdot [\text{PT}]_i^t, \quad (4)$$

The the logits are computed based on the momentum prompt instead of the current prompt token:

$$\boldsymbol{l}^t = g \circ \mathcal{V}\big(\boldsymbol{x}, \{[\text{PT}]_i^{\text{mem}}\}\big), \quad (5)$$

After updating the prompt pool and prompt encoder, the $\alpha^\tau$ needs to be updated with momentum:

$$\begin{aligned}mae &= \text{MAE}(\boldsymbol{l}^t[0:K(t-1)]\cdot\lambda, \boldsymbol{l}^{t-1}\cdot\lambda),\\ \alpha^\tau &= \gamma\cdot\alpha^{\tau-1} + (1-\gamma)\cdot\sigma(mae)\end{aligned} \quad (6)$$

The pseudo code clearly shows that the $\alpha^\tau$ should be updated on each iteration, while memory only needs to be updated once after the current session.

---

**Algorithm 1: Pseudo Code for PEARL**

**Input:** A sequence of $\mathcal{N}$ datasets $\{\mathcal{D}^1,\cdots,\mathcal{D}^\mathcal{N}\}$, prompt pool $\mathcal{P}$, prompt encoder $\mathbf{E}$, a memory space **Memory**;

**Phase 1:**
1: **for** session $t = 1$ **do**
2:    **for** Batch in $\mathcal{D}^1$ **do**
3:      Compute the global prompt $\{[\text{PT}]_i^1\}$ via Eq. (1);
4:      Compute the logits $\boldsymbol{l}^1$ via Eq. (2);
5:      Optimize $\mathcal{P}$ and $\mathbf{E}$ via Eq. (3);
6:    **end for**
7:    Initlize memory by **Memory** $= \{[\text{PT}]_i^1\}$
8: **end for**

**Phase 2:**
9: **for** session $t = 2,\cdots,\mathcal{N}$ **do**
10:    Initialize $\tau = 0$ and $\alpha^\tau = 0.99$;
11:    **for** Batch in $\mathcal{D}^t$ **do**
12:      Compute the global prompt $\{[\text{PT}]_i^t\}$ via Eq. (1);
13:      Obtain the momentum prompt $\{[\text{PT}]_i^{\text{mem}}\}$ via Eq. (4);
14:      Compute the logits $\boldsymbol{l}^t$ via Eq. (5);
15:      Optimize $\mathcal{P}$ and $\mathbf{E}$ via Eq. (3);
16:      Update $\alpha^\tau$ via Eq. (6);
17:      Update $\tau$ by $\tau = \tau + 1$;
18:    **end for**
19:    Update memory by **Memory** $= \{[\text{PT}]_i^{\text{mem}}\}$
20: **end for**

| Method | Task-2 | Task-3 | Task-4 | Task-5 | Task-6 | Task-7 | Task-8 | Task-9 | Task-10 | Average |
|---|---|---|---|---|---|---|---|---|---|---|
| OVOR | 0.993 | 0.996 | 0.995 | 0.998 | 0.998 | 0.999 | 0.998 | 0.999 | 0.999 | 0.997 |
| PEARL w/o NKA | 0.932 | 0.930 | 0.932 | 0.932 | 0.931 | 0.931 | 0.927 | 0.932 | 0.933 | 0.931 |
| PEARL | 0.558 | 0.550 | 0.561 | 0.550 | 0.554 | 0.550 | 0.551 | 0.560 | 0.563 | 0.555 |

Table 1: The cosine similarity before and after prompt updating.

| Method | Metric | IN-A-10 | CIFAR-10 | CUB-10 | IN-R-5 | IN-R-10 | IN-R-20 | Average |
|---|---|---|---|---|---|---|---|---|
| OVOR | $\bar{\mathcal{A}}$ | 48.49 | 86.68 | 78.12 | 76.79 | 75.61 | 73.13 | 73.14 |
|  | $\bar{\mathcal{F}}$ | **7.83** | 5.25 | 8.73 | 4.88 | **5.77** | **6.06** | 6.42 |
| PEARL | $\bar{\mathcal{A}}$ | **67.41** | **91.11** | **92.42** | **82.58** | **81.09** | **79.11** | **82.29** |
|  | $\bar{\mathcal{F}}$ | 9.42 | **4.82** | **3.99** | **4.25** | 5.82 | 6.48 | **5.80** |

Table 2: Comparison with OVOR on six benchmarks.

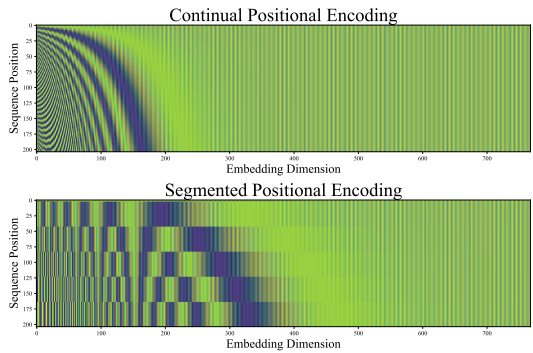

Figure 1: The comparison between CPE and SPE, where SPE uses 5 segments. The sequence length is 200 and the embedding dimension is 768.

## 2 Segmented Positional Encoding VS. Continual Positional Encoding

In this section, we clarify the differences between the proposed segmented positional encoding (SPE) and the continual positional encoding (CPE) (Vaswani et al. 2017). CPE, improves the Transformer's sensitivity to sequence positions by generating a unique vector for each position to indicate its place in the sequence, denoted by the following equation:

$$\begin{aligned} \text{CPE}_{(pos,2j)} &= \sin\left(\frac{pos}{10000^{2j/d}}\right) \\ \text{CPE}_{(pos,2j+1)} &= \cos\left(\frac{pos}{10000^{2j/d}}\right) \end{aligned} \quad (7)$$

In PEARL, the sequence fed into the prompt encoder is a subsequence of the prompt pool, with each subsequence containing knowledge of a specific task. Using CPE would provide positional information at the sequence level, whereas we need it at the task level to identify which task each prompt fragment belongs to. To address this need for task-level positional information, we develop the SPE

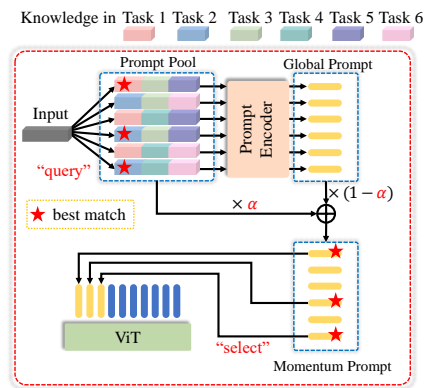

Figure 2: Illustration of the integration of the prompt encoder and the momentum update in "query-select" methods.

method. The equation for SPE is as follows:

$$\begin{aligned} \text{SPE}_{(pos,2j)} &= \sin\left(\frac{\lfloor pos/\frac{\mathcal{M}}{\mathcal{N}} \rfloor}{10000^{2j/d}}\right) \\ \text{SPE}_{(pos,2j+1)} &= \cos\left(\frac{\lfloor pos/\frac{\mathcal{M}}{\mathcal{N}} \rfloor}{10000^{2j/d}}\right) \end{aligned} \quad (8)$$

By comparing Eq. (7) and Eq. (8), it is clear that SPE determines the session number for different positions by dividing the position index by the number of prompts specific to each task. Fig. 1 clearly illustrates this difference. While the CPE assigns a unique position vector to each position, the SPE uses a segmented position vector and maintains consistency within each segment.

## 3 Prompt Encoder on Different Methods

When applying the prompt encoder to an existing method, we follow the framework depicted in Fig. 2. During the "query" phase, no adjustments are made. In the "select"

| Method | $\bar{\mathcal{A}}$ | $\bar{\mathcal{F}}$ | Train Time (ms) | Infer Time (ms) |
|---|---|---|---|---|
| PEARL+DPG | 91.81 | 6.43 | 129.60 | 19.36 |
| PEARL | **94.48** | **4.71** | **70.60** | **13.02** |

Table 3: Impact of instance-level prompt generation on performance.

| Method | CIFAR | CUB | IN-R | IN-A | Omni | VTAB | Average |
|---|---|---|---|---|---|---|---|
| L2P | 8.98 | 11.19 | 11.06 | 19.52 | 16.14 | 33.22 | 16.69 |
| CODA-Prompt | 8.63 | 8.68 | 7.80 | 5.11 | 22.91 | 7.29 | 10.07 |
| RacPAC | 4.05 | 4.74 | 6.45 | 9.05 | 7.13 | 4.19 | 5.94 |
| PEARL | 4.42 | 4.15 | 2.41 | 9.25 | 3.33 | 3.96 | 4.59 |

Table 4: The average forgetting rate ($\bar{\mathcal{F}}$) of multiple methods.

| $H$ | IN-A 10-tasks $\bar{\mathcal{A}}$ | $\mathcal{A}^{\mathcal{N}}$ | VTAB 5-tasks $\bar{\mathcal{A}}$ | $\mathcal{A}^{\mathcal{N}}$ |
|---|---|---|---|---|
| 2 | 66.36 | 57.34 | 95.52 | 92.31 |
| 4 | 66.58 | 57.27 | 95.45 | 92.21 |
| 6 | 66.48 | **57.80** | 95.42 | 92.38 |
| 8 | 66.58 | 57.34 | 95.47 | 92.52 |
| 10 | **66.59** | 57.67 | **95.53** | **92.53** |

Table 5: Ablation study of the length of the prompt token.

| Pool Length | $\bar{\mathcal{A}}$ | $\mathcal{A}^{\mathcal{N}}$ | $\bar{\mathcal{F}}$ |
|---|---|---|---|
| 20 | 66.64 | 57.80 | 11.23 |
| 50 | 66.51 | 57.67 | 10.00 |
| 100 | 67.41 | 57.87 | 9.25 |
| 200 | 66.39 | 58.46 | 9.59 |
| 400 | 66.78 | 58.20 | 9.98 |

Table 6: Ablation study of the length of the prompt token. Red represents the best while blue represents the worst.

phase, we combine the original prompt pool with the encoded global prompt using momentum update and select the prompt based on the best match from the "query" phase. For L2P (Wang et al. 2022b) and DualPrompt (Wang et al. 2022a), the best match is a single prompt, whereas for CODA-Prompt (Smith et al. 2023), it is a distribution across the prompt pool.

The prompt encoder and momentum update in the Fig. 2 correspond to "PE" and "Mom" in main paper Table 2, respectively. When only the prompt encoder is used, the prompt is fed directly from the global prompt. Since the global prompt aggregates all knowledge, the best match obtained through the "query" phase may not accurately reflect the task-specific fitness of the global prompt, leading to performance degradation. Therefore, a knowledge accumulation mechanism is needed for the prompt encoder, and we employ the momentum update. Momentum update allows global knowledge to be integrated into task-related knowledge, improving performance. In the experiment, the momentum weight $\alpha$ is set to a fixed value of 0.99.

## 4  Comparison with State-of-the-art Methods

A task-shared prompt pool eliminates the need for the "query-select" process but constrains the model's ability to adapt to new tasks. To mitigate forgetting, OVOR (Huang, Chen, and Hsu 2024) leverages the numerical stability of prompts to prevent drift, which, in turn, restricts representational diversity. In contrast, our method allocates independent prompts for each task and incorporates the NKA mechanism to promote cross-task diversity. As reported in Table 1, the cosine similarity before and after prompt updates is 0.55 for PEARL, compared to 0.99 for OVOR, underscoring the difference in diversity. As reported in Table 2, across six benchmarks, the accuracy/forgetting values of the proposed PEARL read as 82.2%/5.80%, as compared to 73.14%/6.42% of OVOR, vividly showing the superior of our method. This indicates a positive correlation between prompt diversity and performance, which likely contributes to the observed improvement.

The main distinction between PEARL and existing prompt generation methods is that our prompt tokens are not directly fed to the classifier. Instead, they are momentum-updated using the NKA mechanism, which integrates feedback into the generation process. To our knowledge, this represents the first use of a negative feedback mechanism in incremental learning. Instance-level prompt generation treats each instance as a unique class, expanding the task scale and increasing the risk of forgetting. We modify PEARL following DPaRL (Kim et al. 2025) to implement a instance-level prompt generation. As reported in Table 3, compared to the input-agnostic approach, instance-level prompt generation results in a 2.67% decrease in accuracy and a 1.72% increase in forgetting on CUB-200. Moreover, the instance-level approach incurs significantly higher computational costs, increasing training time by 83.6% and inference time by 48.7% in PEARL. Furthermore, as reported in Table 4, across six benchmarks, the average forgetting rate is 4.59% for PEARL, 5.93% for RacPAC (the previous SOTA), and 10.07% for CODA-P, underscoring PEARL's more effective prompt strategy.

| Init | CIFAR 20-tasks $\bar{A}$ | $\mathcal{A}^{\mathcal{N}}$ | CUB 20-tasks $\bar{A}$ | $\mathcal{A}^{\mathcal{N}}$ | IN-R 40-tasks $\bar{A}$ | $\mathcal{A}^{\mathcal{N}}$ | IN-A 10-tasks $\bar{A}$ | $\mathcal{A}^{\mathcal{N}}$ | Omni 10-tasks $\bar{A}$ | $\mathcal{A}^{\mathcal{N}}$ | VTAB 5-tasks $\bar{A}$ | $\mathcal{A}^{\mathcal{N}}$ |
|---|---|---|---|---|---|---|---|---|---|---|---|---|
| IN21K | **93.64** | **89.02** | 94.48 | 89.65 | 79.54 | 72.33 | 67.41 | 57.87 | **86.87** | 79.68 | 96.52 | **93.02** |
| IN1K | 92.96 | 88.76 | **97.31** | **94.23** | **80.93** | **75.45** | **68.33** | **59.05** | 86.19 | **81.61** | **96.57** | 93.00 |

Table 7: Impact of different initial parameters on performance.

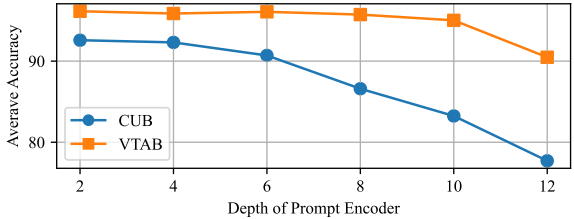

Figure 3: Ablation study on prompt encoder depth.

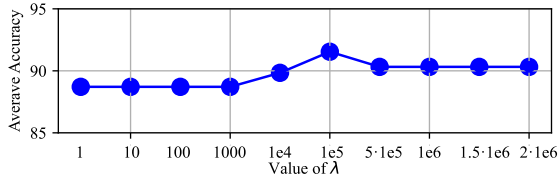

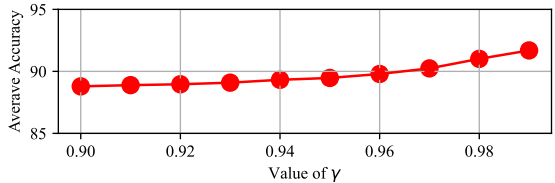

Figure 4: Ablation study on $\lambda$ and $\gamma$.

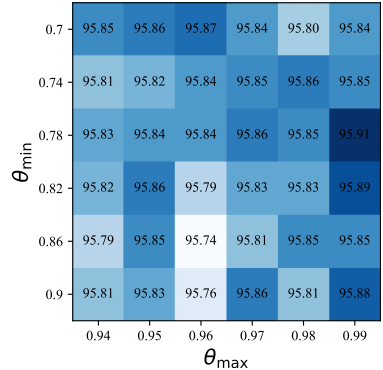

Figure 5: Ablation study on upper and lower bound of the activation function.

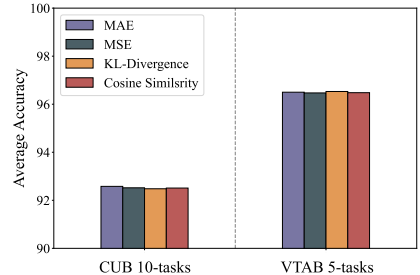

Figure 6: Ablation study on different metrics.

## 5 Further Ablation Study

In this section, we present further ablation study, focusing on hyperparameter selection. For the proposed SPA module, $H$, the length of prompt token and $L$, the depth of prompt encoder are undetermined hyperparameters. Table 5 reports the performance of the model at different length settings. We conduct experiments on ImageNet-A and VTAB, with prompt token length $H$ ranging from $\{2, 4, 6, 8, 10, 12\}$. As shown in Table 5, PEARL maintains relatively consistent performance under different settings. In order to reduce the calculation cost, $H$ is set to 4 in main experiments.

Fig. 3 shows the model performance at different depths. We can see that the model performance deteriorates with increasing depth. This may be due to the fact that more learnable parameters make the model more likely to overfit to the current task. So in the experiment we set $L$ as 2.

We also examine different prompt pool lengths among $\{20, 50, 100, 200, 400\}$. As reported in Table 6, results on the IN-A dataset indicate a maximum difference of 1.02% in accuracy, with 100 being the optimal length. These findings suggest that while prompt token length has minimal impact on performance, prompt pool length is relatively more significant.

We conduct experiments with different initial pre-trained parameters. For PEARL, the IN-1K pre-trained weights achieve an average accuracy improvement of 0.64% over IN-21K. The results are reported in Table 7.

For a sigmoid function $\dot{\sigma}(\cdot)$ with output range of [0,1], to ensure stability during positional momentum updates, we apply a linear mapping to constrain it to a specified range:

$$\sigma(\cdot) = (\theta_{\max} - \theta_{\min}) \cdot \dot{\sigma}(\cdot) + \theta_{\min}. \quad (9)$$

We conduct experiments on the VTAB dataset with $\mathcal{N}$ equals 5. We choose the lower bound $\theta_{\min}$ among $\{0.7, 0.74, 0.78, 0.82, 0.86, 0.90\}$, and the upper bound $\theta_{\max}$ among $\{0.90, 0.92, 0.94, 0.96, 0.98\}$. Fig. 5 shows the performance of the activation function with different upper and lower bound settings. The results demonstrate that the activation

function remains stable and consistent across various parameter combinations, indicating that our NKA method is robust to changes in the activation function's bounds.

For the proposed NKA mechanism, the values of $\gamma$ and $\lambda$ are determined experimentally. $\lambda$ adjusts the divergence between logits to fit the required range for activation functions. For a sigmoid-like activation function, an inappropriate cause the output to depend solely on $\theta_{\min}$ and $\theta_{\max}$, losing its adaptive property. As shown in Fig. 4, an order of 1e4 is appropriate, with $\lambda$ specifically set to 12,500. $\gamma$ determines the rate at which $\alpha$ converges and experiments on different values of $\gamma$ are reported in Fig. 4. We find that performance improves as $\gamma$ increases, so we determine $\gamma$ to be 0.99.

To demonstrate that our NKA method is not dependent on a specific metric, we compare different metrics, including MAE, MSE, KL-divergence, and Cosine Similarity, with the results shown in Fig. 6. The results indicate that varying measurement methods have a limited effect on the final results, confirming the NKA mechanism is a parameterized adaptive mechanism with generalization. The core of this mechanism lies in separating performance measures between different tasks, thereby ensuring stable knowledge transfer.



\end{document}

%% file: m_Intro.tex
\section{Introduction} \label{sec:intro}
% In the realm of computer vision and machine learning, Class-Incremental Learning (CIL)~\cite{iCaRL,CIL-castro2018end,CIL-hou2019learning} has emerged as a crucial paradigm aimed at developing models capable of learning new tasks over time without forgetting previously acquired knowledge. Unlike traditional batch learning, which trains the model on the entire dataset at once, CIL incrementally introduces novel categories, enabling the model to adapt to evolving data distributions and real-world scenarios where new categories may emerge dynamically.
% Incorporating Pre-Trained Models (PTMs) into CIL leverages their rich feature representations, facilitating faster convergence and improved initial performance~\cite{L2P,DualPrompt}. However, when faced with the challenges of dynamic real-world environments, simply fine-tuning PTMs is insufficient. Fine-tuning on new classes can cause ``catastrophic forgetting,'' where the model's performance on previously learned classes significantly degrades due to the overwriting of learned representations~\cite{mccloskey1989catastrophic}.

In the fields of computer vision and machine learning, Class-Incremental Learning (CIL) )~\cite{iCaRL,CIL-castro2018end,CIL-hou2019learning} has become a pivotal paradigm, designed to enable models to acquire new tasks over time without forgetting previously learned information. This approach differs from traditional batch learning, which processes the entire dataset in one go; instead, CIL gradually introduces new categories, allowing the model to adjust to changing data distributions and real-world conditions where new categories may appear dynamically. Integrating Pre-Trained Models (PTMs) into CIL capitalizes on their extensive feature representations, which can speed up convergence and enhance initial performance~\cite{L2P,DualPrompt}. However, in the face of the dynamic challenges posed by real-world environments, merely fine-tuning PTMs proves inadequate. Fine-tuning on new classes may lead to catastrophic forgetting, a phenomenon where a model's performance on old classes deteriorates significantly as new information overwrites existing representations~\cite{mccloskey1989catastrophic}.

\begin{figure}[t]
\centering
    \includegraphics[width=0.48\textwidth]{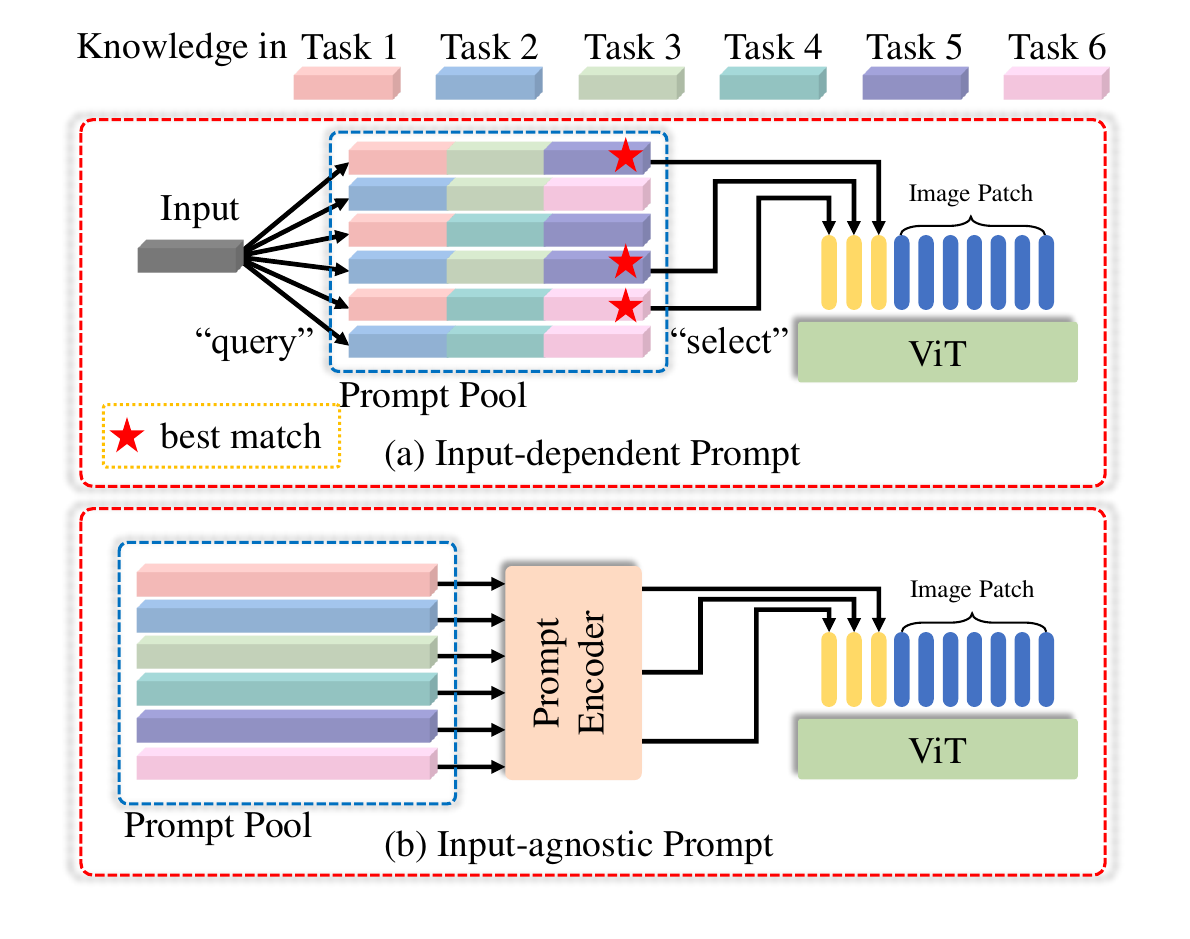}
    \caption{The comparison of (a) input-dependent prompt and (b) input-independent prompt. L2P and DualPrompt following the paradigm in (a), select the best matched prompts, while CODA-Prompt assemble the prompt pool with learnable components during the ``select'' phase.}
     \label{fig:container}
\end{figure}

% To address this challenge, prompt learning has emerged as a novel approach in incremental learning~\cite{L2P,DualPrompt,sprompt,coda}. Prompts, which act as task-specific instructions or context, guide the model in processing input data~\cite{prefixtunning,prompttunning}. In the context of CIL, prompts evolve to accommodate new tasks. 
% To prevent the overwriting of previous knowledge by subsequent tasks in incremental learning, existing methods typically sample in the prompt pool for updating based on input samples~\cite{L2P,DualPrompt,coda,gao2024consistent}. This manner ensures that only a small portion of the prompt pool is updated, thus mitigating catastrophic forgetting. This is described as ``prompt-selection'' by Gao~\etal~\cite{gao2024consistent}, which we further summarize as a ``query-select'' mechanism. For instance, L2P~\cite{L2P} selects the optimal prompts by calculating the similarity between the input and the prompt pool. DualPrompt~\cite{DualPrompt} further divides the prompt into expert prompt and general prompt, and employs a prefix-tuning manner. Coda-Prompt~\cite{coda} addresses the non-differentiability in L2P and DualPrompt by linearly combining the entire prompt pool during the ``select'' phase to produce a fixed-length prompt.

To address the challenges inherent in incremental learning, prompt learning has been introduced as a groundbreaking approach~\cite{L2P,DualPrompt,sprompt,coda}. Prompts serve as task-specific instructions or contextual guides that help the model process input data~\cite{prefixtunning,prompttunning}. In the realm of Class-Incremental Learning (CIL), prompts are dynamically evolved to accommodate new tasks. To prevent the erasure of previous knowledge by subsequent tasks, current strategies typically involve selective sampling from the prompt pool based on input data~\cite{L2P,DualPrompt,coda,gao2024consistent}. This approach ensures that only a subset of the prompt pool is updated at any time, thus reducing the risk of catastrophic forgetting. This is described as ``prompt-selection'' by Gao~\etal~\cite{gao2024consistent}, which we further summarize as a ``query-select" mechanism. For example, L2P~\cite{L2P} identifies optimal prompts by evaluating their similarity to the input and the prompt pool. DualPrompt~\cite{DualPrompt} categorizes prompts into ``expert" and ``general" types and applies them using prefix tuning. Meanwhile, Coda-Prompt~\cite{coda} overcomes the non-differentiability challenges seen in L2P and DualPrompt by linearly combining the entire prompt pool during the selection phase to generate a consistent-length prompt.

% A recent review~\cite{zhou2024continual} has shown that prompt-based methods lag behind other types of approaches in performance. However, Jia~\etal~\cite{ViTPrompt} demonstrate that prompt learning performs well on similar datasets when applied in a supervised learning context. This suggests that while a fixed-length prompt has sufficient expressive power, but the ``query-select'' mechanism  cannot fully exploit its capability. In this paper, we propose a novel theory to explain this phenomenon and provide insights into overcoming the limitations of current prompt-based approaches. We introduce the concept of ``knowledge container'' to explain the shortcomings of current methods. Each prompt in the pool functions as a knowledge container, accumulating knowledge from incoming tasks. As shown in the Fig.~\ref{fig:container}, the ``query-select'' mechanism often mixes knowledge from multiple tasks in an unstructured manner. This unstructured mixing leads to inconsistent knowledge representations and difficulties in maintaining coherent task-specific information. The absence of a structured approach to manage and preserve task-specific knowledge undermines the effectiveness of existing methods.

A recent review in~\cite{zhou2024continual} reveals that prompt-based methods generally underperform compared to other approaches. Conversely, Jia \etal. demonstrate that prompt learning is effective on comparable datasets within a supervised learning context~\cite{ViTPrompt}. This indicates that although fixed-length prompts possess adequate expressive potential, the ``query-select" mechanism fails to fully harness this capability. In this paper, we propose a novel theory to elucidate the observed phenomenon and offer insights into surmounting the limitations of existing prompt-based methods. We introduce the concept of a ``knowledge container" to detail the shortcomings of these approaches. Each prompt in the pool serves as a knowledge container, accumulating insights from incoming tasks. As illustrated in Fig.~\ref{fig:container}, the ``query-select" mechanism frequently amalgamates knowledge from various tasks in a disorganized fashion. This haphazard mixing results in inconsistent knowledge representations and challenges in preserving coherent, task-specific information. The lack of a systematic method to manage and safeguard task-specific knowledge significantly detracts from the effectiveness of existing approaches.

% To address the limitations of the ``query-select'' mechanism, we propose creating an input-agnostic prompt that is suitable for all instances within the same session.	
% Given the temporal nature of incremental learning, we model the prompt adaptation process as a sequential problem, enabling steady incremental learning by gradually capturing correlations between tasks. We refer to this input-agnostic prompt as \textbf{Sequential Prompt Adaptation (SPA)}. During each session, the prompt pool is encoded by {\color{red}a session-sharing prompt encoder} {\color{blue}(appropriate?)} to produce a global prompt. By freezing certain parameters in the prompt pool, we implement independent knowledge containers for each task while maintaining uniform knowledge through the prompt encoder.
% Drawing inspiration from relative positional encoding~\cite{dai2019transformer}, we propose the segmented positional encoding to ensure consistency within each segment of the prompt pool.

To overcome the limitations of the ``query-select" mechanism, we propose the creation of an input-agnostic prompt suitable for all instances within a single session. Considering the temporal dynamics of incremental learning, we conceptualize the prompt adaptation process as a sequential problem, facilitating steady incremental learning by progressively capturing task correlations. We refer to this approach as \textbf{Sequential Prompt Adaptation} (\textbf{SPA}). During each session, a session-sharing prompt encoder processes the prompt pool to generate a global prompt. Following~\cite{coda}, we create independent knowledge containers for each task by freezing specific parameters in the prompt pool, while ensuring uniform knowledge representation through the prompt encoder. Further, we introduce a segmented positional encoding to maintain consistency within each segment of the prompt pool.

% Since we eliminate the ``query-select'' mechanism and introduce {\color{red}a session-sharing prompt encoder}, the model may easily overfit to the current task. To mitigate this risk, we introduce the \textbf{Negative-feedback Knowledge Accumulation (NKA)} mechanism. In this mechanism, the prompt is updated using a momentum-based approach, where the momentum weight is derived from the model's output, which is influenced by the prompt itself. The momentum weight is determined by the divergence between the current output and the previous output. A low divergence indicates that old knowledge has been well retained, allowing the prompt to acquire more new knowledge (\ie reduce weight). Conversely, a high divergence suggests insufficient retention of old knowledge, necessitating greater emphasis on retaining the old parameters (\ie enlarge weight). In essence, this mechanism ensures that the model's ability to learn new tasks is contingent upon its retention of old knowledge, aligning closely with the goals of incremental learning. 

By eliminating the ``query-select" mechanism and introducing a session-sharing prompt encoder, our model risks overfitting to the current task. To counteract this, we introduce the \textbf{Negative-feedback Knowledge Accumulation} (\textbf{NKA}) mechanism. This approach updates the prompt using a momentum-based method, where the momentum weight is influenced by the model's output, which in turn is affected by the prompt itself. The weight is adjusted based on the divergence between current and previous outputs. Low divergence, indicating well-retained old knowledge, allows the prompt to integrate more new knowledge (\ie, reduce weight). Conversely, high divergence, showing poor retention of old knowledge, requires increased focus on preserving previous parameters (\ie, increase weight). Essentially, this mechanism ensures knowledge retention as a prerequisite for learning new tasks. Additionally, the NKA mechanism helps to reveals potential correlations between tasks, thereby enhancing knowledge accumulation.

Our contributions include:
\begin{itemize}
    \item We propose a novel CIL framework called \textbf{PEARL}, where a prompt encoder generates uniform prompts infused with global knowledge and accumulates knowledge through a momentum-based update strategy driven by negative feedback regulation. 
    % PEARL demonstrates that prompting-based approaches can achieve exceptional performance in PTM-based CIL.
    \item We introduce the \textbf{SPA} module, which enables a global prompt to simultaneously encapsulate knowledge from different tasks, overcoming the shortcomings of current ``query-select'' mechanism.    
    \item The proposed \textbf{NKA} mechanism effectively implements an adaptive momentum update, achieving efficient knowledge accumulation by leveraging inherent data correlations.
    \item Through extensive experiments, we demonstrate that our method achieves state-of-the-art performance, surpassing the second-best results by an average of 2.24\% in accuracy across six benchmarks.
\end{itemize}

%% file: m_Related.tex
\section{Related Work} \label{sec:related}
\subsection{Class-Incremental Learning}

CIL is one of the research hotspots in machine learning. Its main challenge is ``catastrophic forgetting'', which occurs when the model overfits the current task and loses knowledge from previous tasks. According to~\cite{masana2022class}, there are three main technical approaches for existing CIL researches. Rehearsal-based approaches reduce forgetting by either retaining a limited set of representative samples or generating pseudo-samples~\cite{rehear-rebuffi2017icarl,rehear-shin2017continual,rehear-xiang2019incremental,rehear-ostapenko2019learning}. Methods based on regularization consider to impose constraints on the representation or weight of the model~\cite{regu-jung2016less,regu-kirkpatrick2017overcoming,regu-li2017learning,regu-aljundi2018memory,regu-chaudhry2018riemannian}, and usually use knowledge distillation technology~\cite{hinton2015distilling} to enhance the memory ability of the model. The bias-correction approach aims to solve the domain shift problem by aligning feature distribution between different tasks to alleviate overfitting when the model is faced with new tasks~\cite{bic-castro2018end,bic-hou2019learning,bic-wu2019large}.

\begin{figure*}[t]
\centering
    \includegraphics[width=0.9\textwidth]{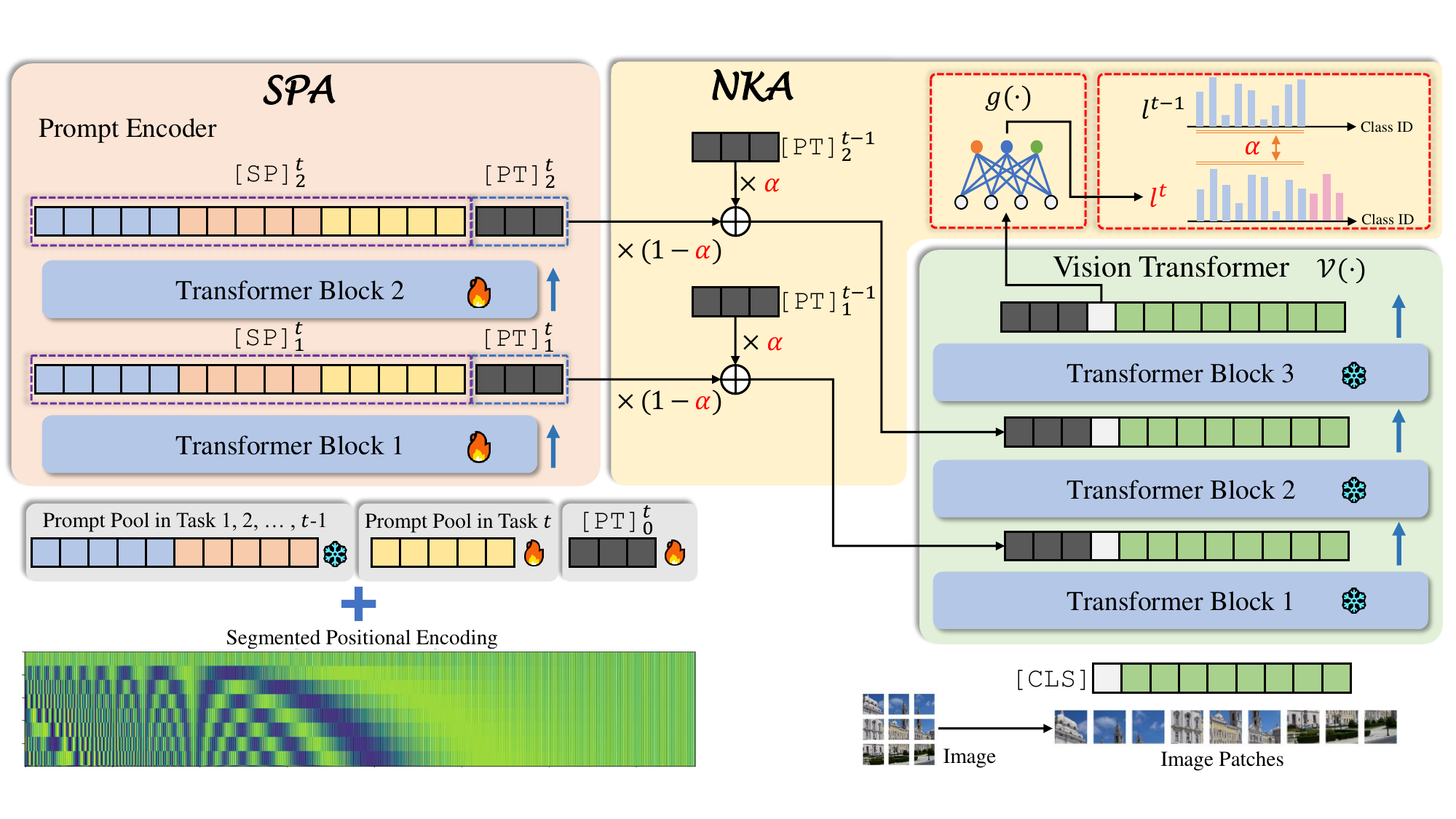}
    \caption{The illustration of the proposed PEARL. The ViT consists of 12 blocks, with a $L$-layer prompt encoder added to the last few blocks of the ViT. The ViT is frozen during training, while the prompt encoder and prompt pool remain learnable.}
     \label{fig:mainfigure}
\end{figure*}

In recent years, with the rise of PTMs, many researches focus on PTM-based CIL. According to the recent review research~\cite{zhou2024continual}, the existing methods can be divided into three categories. The prompt-based methods focus on prompt learning in CIL~\cite{L2P,DualPrompt,coda,ovor,dparl}. Leveraging the robust representational capabilities of PTMs, researchers have demonstrated that effective continuous learning on downstream tasks can be achieved by fine-tuning a prompt. Similarly, the representation-based approach focuses on rich representations of PTMs and improve their generalization in CIL through regularization and metric learning~\cite{SimpleCIL,RanPAC,EASE}. In addition, the approach based on model-mixture has also received attention~\cite{HiDePrompt,ESN}. The key of model-mixture is to assign corresponding expert models to different tasks through ensemble learning, so as to improve the overall capability of the model. Zhou~\etal~\cite{zhou2024continual}, through comprehensive comparison, find that the prompt-based methods underperform compared to the other two mainstream approaches. Through in-depth analysis, we propose a ``knowledge container'' theory to explain this phenomenon, and designs a novel prompt method based on this theory, which reaches the state-of-the-art result.
\subsection{Prompt Learning for Pre-Trained Models}
With the success of PTMs in the field of natural language processing (NLP), similar techniques have been introduced to computer vision (CV) tasks. 
Prompt Tuning~\cite{prompttunning,zhang2024causal} is a method of tuning models by adding learnable prompt tokens before input data. Prompt Tuning helps models perform better by aligning distribution of downstream tasks and pre-trained data. Prefix Tuning~\cite{prefixtunning} establishes a flexible attention mechanism, where learnable prompts are appended to the attention parameters. Additionally, researchers have also developed several other fine-tuning methods (\eg BitFit~\cite{zaken2021bitfit} and LoRA~\cite{hu2021lora}) to further improve the efficiency of tunning.

%% file: m_method.tex
\section{Methodology} \label{sec:method}
\subsection{Problem Formulation}
In CIL, training data appears in the form of data stream and each session in the stream contains a task. A data stream with $\mathcal{N}$ sessions can be refered to as:$\{\mathcal{D}^{1},\cdots,\mathcal{D}^{\mathcal{N}}\}$, where $\mathcal{D}^{t}=\{(x_i,y_i)\}_{0<t\leq \mathcal{N}}$ is the training set for the $t$-th session and $y_i$ belongs to the class set $\mathcal{C}^{t}$. Each task contains the same number of categories, \ie $\vert \mathcal{C}^{t} \vert = K,~\mathrm{for}~t = 1,2,\cdots, \mathcal{N}$. The settings of CIL require that datasets at different sessions cannot share class labels, \ie $\mathcal{C}^{i} \cap \mathcal{C}^{j}=\emptyset,~\mathrm{for}~\mathrm{any}~i \neq j$. At the same time, the model is required to retain the memory of all previous tasks during testing, so the test set at the $t$-th session needs to contain all previous labels,~\ie $C_{\mathrm{test}}^t=\mathcal{C}^{0} \cup \mathcal{C}^{1}\cdots \cup \mathcal{C}^{t}$.

\subsection{Overview on PEARL }

% Based on existing research on PTM-based CIL, our model applies models pre-trained on large-scale data sets and is expected to generalize to downstream tasks. In particular, for input $\Vec{x}$, the model can be denoted as: $\mathcal{V}(\Vec{x})\in \mathbb{R}^{d}$, where $d$ is the dimention of the features. At the $t$-th session, the predicti on logit $\Vec{l}^{t}$ is obtained using a prototype-based classification header:

Recognizing the limitations of existing ``query-select'' methods, we propose the \textbf{PEARL} to build an input-agnostic prompt. PEARL consists of two components: the \textbf{SPA} module and the \textbf{NKA} mechanism. In SPA, we model the prompt learning process as a sequential problem, leveraging the inherent temporal dynamics of incremental learning. Following CODA-Prompt~\cite{coda}, only a proportion of the prompt pool is updated in the corresponding session to facilitate explicit knowledge management. The prompt encoder $\mathbf{E}$ and the 
prompt pool $\mathcal{P}$ are defined as follows:
\begin{equation}\label{eq1}
\begin{split}
    \mathbf{E} &= \{\mathbf{B}_1, \mathbf{B}_2,\cdots, \mathbf{B}_{{L}}\},
    \\
    \mathcal{P} &= \{\mathbf{P}_1, \mathbf{P}_2,\cdots, \mathbf{P}_{\mathcal{M}} \},
\end{split}
\end{equation}
where, $\mathbf{B}_i$ represent the $i$-th block in the prompt encoder, $L$ represents the depth of the prompt encoder and $\mathcal{P}$ contains learnable prompts with the number of $\mathcal{M}$.

We choose ViT~\cite{ViT} as the backbone model $\mathcal{V}(\cdot)\in\mathbb{R}^d$ and implement a classification head $g(\cdot)$ which contains no trainable parameters and updated following the RanPAC manner~\cite{RanPAC}. Inspired by the $\texttt{[CLS]}$ token in ViT, we design a prompt token $\texttt{[PT]}$ to obtain a fixed-length prompt after aggregating the prompt pool. During session $t$, the model can derive a prompt token $\texttt{[PT]}_i^t\in \mathbb{R}^{H\times d}$ with length $H$ from the $i$-th block of $\mathbf{E}(\cdot)$, generating diverse representations and forming a set of prompts: $\{\texttt{[PT]}_i^t\}_{1\leq i\leq L}\in \mathbb{R}^{L\times H\times d}$. For a certain instance $\Vec{x}$, The prediction logit $l^t$ is computed by:
\begin{equation}
    l^t = g \circ \mathcal{V}\big(\Vec{x},\{\texttt{[PT]}_i^{\mathrm{mem}}\}\big),
\end{equation}
where $\texttt{[PT]}_i^{\mathrm{mem}}$ will be defined by Eq.~\eqref{eq:mem}.
% Following the settings of~\cite{L2P,DualPrompt}, we use the pre-trained parameter of ViT, so $\mathcal{V}(\cdot)$

From the perspective of knowledge containers, ${\texttt{[PT]}^t_i}$ aggregates previous knowledge from the prompt pool. However, this knowledge serves primarily as a good initialization and is prone to overfitting.
% Inspired by the negative feedback regulation, we design the NKA mechanism where $\{\texttt{[PT]}_i\}$ is updated by the momentum manner and the momentum weight is decided by the capability of retaining previous knowledge.
The proposed NKA mechanism address this problem by introducing a momentum update strategy:
\begin{equation}\label{eq:mem}
    \texttt{[PT]}_{i}^{\mathrm{mem}} = \alpha^{\tau}\cdot\texttt{[PT]}_{i}^{t-1}+(1-\alpha^{\tau})\cdot\texttt{[PT]}_{i}^{t},
\end{equation}
where the momentum prompt $\texttt{[PT]}_{i}^{\mathrm{mem}}$ will be sent into the backbone and $\alpha^{\tau}$ represents the momentum weight which is obtained through the negative feedback regulation.
By mixing knowledge from different sessions, the NKA mechanism ensures the stability of old knowledge while also acquiring new knowledge.	

The backbone $\mathcal{V}(\cdot)$ remains frozen during incremental learning, and the primary objective is to identify the optimal prompt encoder and prompt pool:
\begin{equation}
    \mathbf{E}^{*},\mathcal{P}^{*} = \underset{\mathbf{E},\mathcal{P}}{{\arg\max}} \, \mathbb{E}_{(\Vec{x},y)\sim\mathcal{D}^{t}}\mathbb{I}(y\neq l^t)).
\end{equation}
The pseudo code is provided in the supplementary material.
% an adaptive momentum update method that can dynamically adjust momentum weights according to current utility, thus achieving stability during knowledge transfer.

% design a momentum prompting method for learning new knowledge for downstream tasks. Typical prompt-based methods in CIL can be summarized as a "query-selection" mechanism that samples a prompt pool based on model inputs. This fine-tuning paradigm poses two problems. (1) Fine-tuning dependent on input is difficult to adapt to the distribution differences between different sessions in incremental learning. Due to this "query-selection" mechanism, prompt pool updates may be delayed, so there is no guarantee that all prompts can be synchronized. This intermittent update approach results in prompt-based methods being at a disadvantage in PTM-based CIL research. (2) The prompt pool mixes the knowledge learned at different sessions, without being able to distinguish it explicitly, which can easily lead to catastrophic forgetting.

% To solve the above problems, we propose an instance-independent fine-tuning method, momentum fine-tuning. To gain global knowledge, our MoPo method models CIL as a time series task, allowing knowledge to be learned continuously at different sessions. 

% We first introduce the Sequential Prompt Adaptation (SPA), then the Negative-Feedback Knowledge Accumulation (NKA), and then we summarize the overall optimization progress of the model.
\subsection{Sequential Prompt Adaptation }
The primary challenge in constructing the input-agnostic prompt is establishing cross-task information interaction.	
This is because CIL needs to be backward-compatible: new tasks must build upon previous knowledge rather than requiring a complete rebuild.
% Since ViT requires the input to the prompt always to be a sequence of equal length, prompt-based methods need to encode the prompt pool.
% During this process, cross-task information interaction ensures that previous memories are not lost. 
% ~\cite{wang2024comprehensive,zhou2024class}

Existing prompt-based methods~\cite{L2P,DualPrompt,coda} achieve this interaction through a ``prompt-input-prompt'' link. However, the input-agnostic prompt cannot establish such a link. We propose to omit the intermediate link and enable direct interaction between prompts in the form of a sequence.
We utilize the Transformer architecture~\cite{transformer} to capture the sequential relationship. The prompt encoder, defined in Eq.~\eqref{eq1}, consists of $L$ blocks, with the output function after the $i$-th block denoted as: 
\begin{equation}
    \mathcal{E}_i(\cdot) = \mathbf{B}_1 \circ \mathbf{B}_2 \circ \cdots \circ \mathbf{B}_i(\cdot),
\end{equation}
and the subset of the prompt pool is denoted as:
\begin{equation}
    \mathcal{P}[1:k] = \{\mathbf{P}_1, \mathbf{P}_2,\cdots, \mathbf{P}_{k} \}. 
\end{equation}
During the $t$-th session, only $\mathcal{P}[1+\frac{\mathcal{M}}{\mathcal{N}}\times (t-1):\frac{\mathcal{M}}{\mathcal{N}}\times t]$ are learnable while other prompts are frozen to keep previous knowledge.
% Since $\{\texttt{[PT]}_i\}$ is constantly updated in the process of incremental learning, it is necessary to mark $\{\texttt{[PT]}_i\}$ at different sessions by superscript (\eg $\texttt{[PT]}^1_1, \texttt{[PT]}^2_1$). 
% Observing the temporal correlation in incremental learning, we suggest constructing the prompt encoding process as a sequence of tasks over time. By encoding the input prompt sequence, we obtain an output sequence. By adding a set of learnable vectors to the initial position in the input sequence, we can achieve a fixed-size $\texttt{[PT]}$ token from a varying sequence length.
% We propose a prompt encoder $\mathcal{M}=\{B_1,B_2,\cdots,B_L\}$ consisting of $L$ transformer blocks to achieve hierarchical prompt encoding. In standard transformers, positional encoding assigns a unique encoding to each position in the sequence. However, in the context of continual incremental learning (CIL), prompts at the same time step should share the same encoding information. 
The process of prompt encoding can be formulated as below:
\begin{equation}
\begin{split}    
    &\texttt{[PT]}_1^t, \texttt{[SP]}_1^t = \mathcal{E}_1\big(\mathrm{ConCat}(\texttt{[PT]}_{0}^t,\mathcal{P}[1:\frac{\mathcal{M}}{\mathcal{N}}\times t])\big),\\
    &\texttt{[PT]}_2^t, \texttt{[SP]}_2^t = \mathcal{E}_2 \big(\mathrm{ConCat}(\texttt{[PT]}_{1}^t,\texttt{[SP]}_{1}^t)\big),\\
    &\quad\quad\quad\quad\quad\quad \vdots  \\
    &\texttt{[PT]}_i^t, \texttt{[SP]}_i^t = \mathcal{E}_i \big(\mathrm{ConCat}(\texttt{[PT]}_{i-1}^t,\texttt{[SP]}_{i-1}^t)\big),
\end{split}
\end{equation}
where $\texttt{[SP]}_i^t$ is short for sequential prompts after the $i$-th block and is the intermediate variable during encoding. The data flow can be seen in Fig.~\ref{fig:mainfigure}.
We adopt the prefix-tunning manner~\cite{prefixtunning}, and the $\texttt{[PT]}_{i}^t$ can be further embedded as learnable prefixes: $\Vec{p}_K ,\Vec{p}_V \in \mathbb{R}^{L\times H\times d}$. The learnable prefixes are attached in the Multi-head Self-attention (MSA):
\begin{equation}
    f_{\mathrm{prefix}} = \mathrm{MSA}\big(\Vec{h}_Q,\mathrm{ConCat}(\Vec{p}_K,\Vec{h}_K), \mathrm{ConCat}(\Vec{p}_V,\Vec{h}_V)\big),
\end{equation}
where $\Vec{h}_Q, \Vec{h}_K, \Vec{h}_V$ are attention parameters.

% To achieve this, we propose a segmented positional encoding approach. 
Instead of encoding each position individually, SPA encodes the input sequence according to the task number. To achieve this, we introduce a segmented positional encoding (SPE): 
\begin{equation}
\begin{split}    
    \mathrm{SPE}_{(pos,2j)}\quad &= \sin\left(\frac{\lfloor pos / \frac{\mathcal{M}}{\mathcal{N}}\rfloor}{10000^{2j} / d}\right), \\
    \mathrm{SPE}_{(pos,2j+1)} &= \cos\left(\frac{\lfloor pos / \frac{\mathcal{M}}{\mathcal{N}}\rfloor}{10000^{2j} / d}\right),
\end{split}
\end{equation}
where $pos, j$ indicates the location and $d$ denotes the dimension of feature vectors. For $pos \in \{H+1,H+\frac{\mathcal{M}}{\mathcal{N}}\}$, the $\texttt{[SP]}_i^t$ share the same positional encoding because they represent knowledge from the same task.
SPE incorporates session information into the sequential prompts, thereby enhancing the model's ability to learn and retain task-specific knowledge.	
Additional details and visualization of SPE is available in the supplementary material.
\subsection{Negative-feedback Knowledge Accumulation}
\begin{figure}[t]
\centering
    \includegraphics[width=0.45\textwidth]{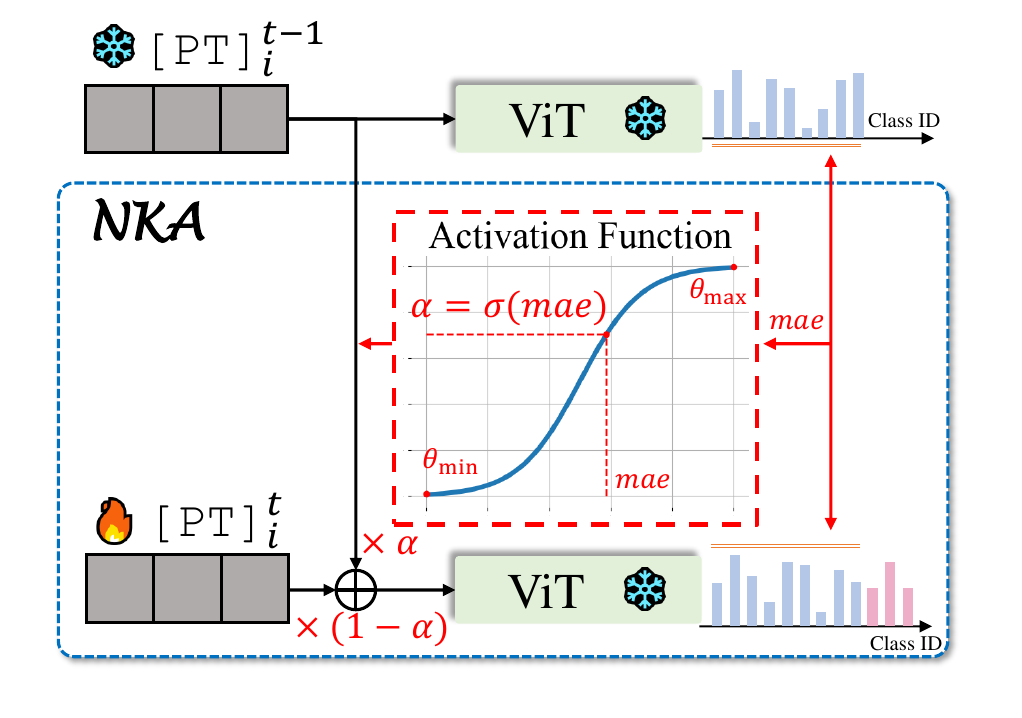}
    \caption{The illustration of the proposed NKA mechanism. A low $mae$ indicates good knowledge retention, enabling the model to focus more on the new task. }
     \label{fig:NKA}
\end{figure}

Inspired by the negative feedback regulation, we propose a knowledge accumulation mechanism. Specifically, we assess knowledge retention by computing the divergence between the current logits and those from the previous task. This divergence serves as a feedback signal used to dynamically adjust the weights of the prior prompt token (\ie $\{\texttt{[PT]}^{t-1}_i\}$) and the current prompt token (\ie $\{\texttt{[PT]}^{t}_i\}$). The flow of the NKA mechanism is shown in Fig.~\ref{fig:NKA}. Given two logits $\Vec{l}^{t}$ and $\Vec{l}^{t-1}$, the divergence is computed as Mean Absolute Error (MAE):
\begin{equation}
    mae = \mathrm{MAE} ( \Vec{l}^{t}[0:{K}(t-1)]\cdot\lambda,\Vec{l}^{t-1}\cdot\lambda ),
\end{equation}
where $\lambda$ servevs as a scale factor.
% \begin{equation}
%     sim=\frac{\Vec{l}^{t}[0:{K}(t-1)]\cdot\Vec{l}^{t-1}}{\vert\vert\Vec{l}^{t}[0:K(t-1)]\vert\vert \cdot \vert\vert\Vec{l}^{t-1}\vert\vert}.
% \end{equation}
The computation involves only the first ${K}(t-1)$ terms of the current logits, as the decision space is expanding and the last $K$ terms represent new knowledge rather than previous knowledge. The momentum weight $\alpha^{\tau}$ is further computed by:
\begin{equation}\label{eq:backward}
    \alpha^{\tau} = \gamma\cdot\alpha^{\tau-1}+(1-\gamma)\cdot\sigma(mae),
\end{equation}
where $\tau$ is the iteration number and $\sigma(\cdot)$ is the sigmoid-like activation function with an upper bound $\theta_{\mathrm{max}}$ and a lower bound $\theta_{\mathrm{min}}$.  We employ a momentum update for $\alpha^{\tau}$ to ensure numerical stability and prevent fluctuations that could lead to drastic changes in $\{\texttt{[PT]}^{t-1}_i\}$.
When updating $\{\texttt{[PT]}^{\mathrm{mem}}_i\}$, the momentum update is denoted by Eq.~\eqref{eq:mem}.

Eq.~\eqref{eq:mem} and Eq.~\eqref{eq:backward} defines the feedback process of negative-feedback regulation. Further, the forward process of NKA is denoted as follows:
\begin{equation}
\begin{split}    
    &\Vec{l}^{t-1}=g \circ \mathcal{V}(\Vec{x},\{\texttt{[PT]}_{i}^{t-1}\}),\\
    &\Vec{l}^{t}\quad=g \circ \mathcal{V}(\Vec{x},\{\texttt{[PT]}_{i}^{\mathrm{mem}}\}).
\end{split}
\end{equation}
The current momentum token $\{\texttt{[PT]}_{i}^{\mathrm{mem}}\}$ is stored in memory and serves as $\{\texttt{[PT]}^{t}_i\}$  during the $(t+1)$-th session. Since PEARL is an input-agnostic method, it requires only additional memory space of size $L\times H\times d$, avoiding inference costs for computing  $\{\texttt{[PT]}_{i}^{t-1}\}$. 

In optimization, the Cross-Entropy loss is used during the $t$-th session:
\begin{equation}
    \mathcal{L}_{\mathrm{cls}}=\mathbb{E}_{(\Vec{x},y)\sim\mathbf{D}^{t}} \mathrm{CE}(\Vec{l}^{t},y).
\end{equation}
Our method does not require any knowledge retention loss (\eg Knowledge Distillation), because knowledge retention is accounted for by the adaptive weighting factor $\alpha^{\tau}$. Our model begins to fit new tasks only if there is no forgetting of old tasks; otherwise, the prompt token will continuously backtrack. Consequently, PEARL elevates prompt-based methods to state-of-the-art performance, challenging existing beliefs about their efficacy for PTM-based CIL.
% \subsection{Full Optimization}

% \begin{equation}

% \end{equation}

%% file: m_Expt.tex
\section{Experiments} \label{sec:expt}

\subsection{Datasets and Implementation Details}

\input{tables/table1}

\noindent\textbf{Datasets. }In order to comprehensively examine the model performance, we follow~\cite{zhou2024class} and conduct experiments on six datasets including CIFAR00~\cite{cifar}, CUB200~\cite{cub}, ImageNet-R~\cite{imagenetr}, ImageNet-A~\cite{imageneta}, Omnibenchmark~\cite{omni} and VTAB~\cite{vtab}. There are 50 classes in VTAB, 100 classes in CIFAR-100, 200 classes in CUB200, ImageNet-R and ImageNet-A and 300 classes in Omnibenchmark.

% The details of the datasets are listed below: {\color{red}page control}
% \begin{itemize}
%     \item \noindent\textbf{CIFAR-100} contains 100 categories
%     \item \noindent\textbf{CUB200} contains 100 categories
%     \item \noindent\textbf{ImageNet-R} contains 100 categories
%     \item \noindent\textbf{ImageNet-A} contains 100 categories
%     \item \noindent\textbf{Omnibenchmark} contains 100 categories
%     \item \noindent\textbf{VTAB} contains 100 categories

% \end{itemize}

\noindent\textbf{Evaluation metrics.} For an incremental learning task with $\mathcal{N}$ sessions in total, the classification accuracy of the model on the $t$-th session is denoted as $\mathcal{A}^t$. Followed by~\cite{zhou2024class}, we adopt two evaluation metrics: the average accuracy $\bar{\mathcal{A}}=\frac{1}{\mathcal{N}}\sum_{t=1}^{\mathcal{N}}\mathcal{A}^t$ and the final accuracy $\mathcal{A}^{\mathcal{N}}$. 
% Since all comparison methods in PTM-based CIL utilize the same backbone network and pre-trained parameters, the accuracy provides a reliable measure of performance. 

\noindent\textbf{Implementation details.} Following the experiment settings of~\cite{L2P,ESN,EASE,zhou2024class}, We choose ViT as the backbone initialized with \textbf{ViT-B/16-IN21K} and \textbf{ViT-B/16-IN1K} parameters. $L$ and $H$ equals 2 and 4, respectively and the length of prompt pool is 100. In NKA mechanism, the initial $\alpha^0$ is set as 0.99 and $\lambda$ equals 12500 during the training process. The upper and lower bounds of $\sigma(\cdot)$ are set as 0.999 and 0.7, respectively. We train the model with SGD optimizer and cosine annealing with epoch as 10 and batchsize as 32. Our results are the average of three random runs and conducted with PyTorch\cite{paszke2019pytorch} and PILOT\cite{sun2023pilot}. All experiments are conducted on one RTX 4090. 

\subsection{Comparison with State-of-the-art Methods}

The compared methods include prompt-based (\eg L2P~\cite{L2P}, DualPrompt~\cite{DualPrompt}, CODA-Prompt~\cite{coda}), representation-based (\eg SimpleCIL~\cite{SimpleCIL}, ADAM~\cite{SimpleCIL}, RanPAC~\cite{RanPAC}, EASE~\cite{EASE}) and model mixture-based (\eg HiDe-Prompt~\cite{HiDePrompt}, ESN~\cite{ESN}).

As reported in Table~\ref{tab:1}, PEARL achieves the best performance among all six benchmarks.
% , outperforming the second best methods by an average of {\color{red}5\%} and {\color{red}6\%} on $\bar{\mathcal{A}}$ and $\mathcal{A}^{\mathcal{N}}$, respectively. 
The experiments span various sequential lengths, and our method performs well in a variety of settings, demonstrating its superiority. Compared to other prompt-based methods (\ie L2P, DualPrompt and CODA-Prompt), our method demonstrates a significant advantage, with average improvements of {13.58\%} and {15.02\%} on $\bar{\mathcal{A}}$ and $\mathcal{A}^{\mathcal{N}}$. This improvement is attributed to the integration of the proposed SPA module and NKA mechanism, which will be further analyzed in the ablation study. 
Compared to RanPAC, the second-best method, PEARL achieves an average improvement of {2.24\%} and {1.65\%} on $\bar{\mathcal{A}}$ and $\mathcal{A}^{\mathcal{N}}$, respectively.  
In RanPAC, the model updates only at $t=1$ and adjusts the classification head based on features from later sessions, limiting its ability to effectively learn from subsequent tasks. In contrast, PEARL performs continuous updates across all sequential tasks, ensuring that the latest knowledge is consistently learned.

\subsection{Ablation Study}
\input{tables/table5}
\textbf{Effect of SPA module.} 
To verify the generality of the proposed SPA module, we add the prompt encoder to other prompt-based methods and make comparisons. As reported in Table~\ref{tab:prompt_ablation}, the model benefits only when both the prompt encoder and momentum update are combined; using the prompt encoder alone leads to significant drawbacks. This is because the prompt encoder mixes task-specific knowledge, making the ``query-select'' mechanism ineffective.	DualPrompt partially maintained its performance due to its unique ``general-expert'' prompt design, whereas L2P and CODA-Prompt experienced significant degradation.	However, a simple fixed-weight momentum update addresses this issue by enabling smooth knowledge accumulation. We conclude that implementing an input-agnostic prompt effectively requires both a global encoder and a momentum update strategy; relying on either alone is insufficient. 
Table~\ref{tab:posi} reports the impact of different positional encodings. We compare the continual positional encoding~\cite{transformer} with the proposed segmented positional encoding. The experiments demonstrate that segmented positional outperforms on both ImageNet-A and VTAB.
Further details are provided in the supplementary material.	

% we can find that introducing sequential relationships between prompts can improve performance effectively. However, due to the limitations of the ``query select'' mechanism, the encoder cannot work during the ``query'' phase, which limits the performance of the SPA module. Details on adding the prompt encoder to existing methods will be reported in the supplementary material.

% To make a fair comparison with the existing prompt-based method, we introduce the prompt encoder and keep its number of parameters consistent with PEARL. As reported in Table~\ref{tab:prompt_ablation}, we can find that introducing more learnable parameters can improve performance, but due to the limitation of ``query-select'' mechanism, increasing the number of parameters does not bring significant performance improvement. Our PEARL method, however, brings a more efficient way to fine-tune PTM-based CIL.

% We implement two popular prompt methods which are prompt-tunning and prefix-tunning. In order to evaluate the performance gain brought by SPA as fairly as possible, the momentum update weight here is fixed at 0.9. We compare the effect of different prompt manners and add prompt at the last $L$ blocks of ViT, with $L$ ranging from \{1,2,3,4\}. As reported in Table~\ref{tab:prompt_ablation}, prefix-tunning outperforms prompt-tunning in general performance, and the accuracy increases with the number of layers. 
\input{tables/table4}

\noindent\textbf{Effect of NKA mechanism.} 
We compare the NKA update with a fixed-weight momentum update. As shown in Table~\ref{tab:nka}, when $\alpha$ falls below 0.9, performance deteriorates rapidly, resulting in the complete failure of the fixed-weight momentum update. This suggests that a fixed $\alpha$ makes the model highly sensitive to the initial value of $\alpha^0$. Meanwhile, when $\alpha$ is updated based on the NKA mechanism, the model achieves better results across different initial conditions, with $\bar{\mathcal{A}}$ and $\mathcal{A}^{\mathcal{N}}$ improved by an average of 14.37\% and 18.09\%, respectively. 
Additionally, we perform further analysis of the update process of $\alpha^{\tau}$, which will be discussed in the following part.

\subsection{Further Analysis}\label{FurtherAnalysis}
\input{tables/table2}

\begin{figure*}[t]
\centering
    \includegraphics[width=0.95\textwidth]{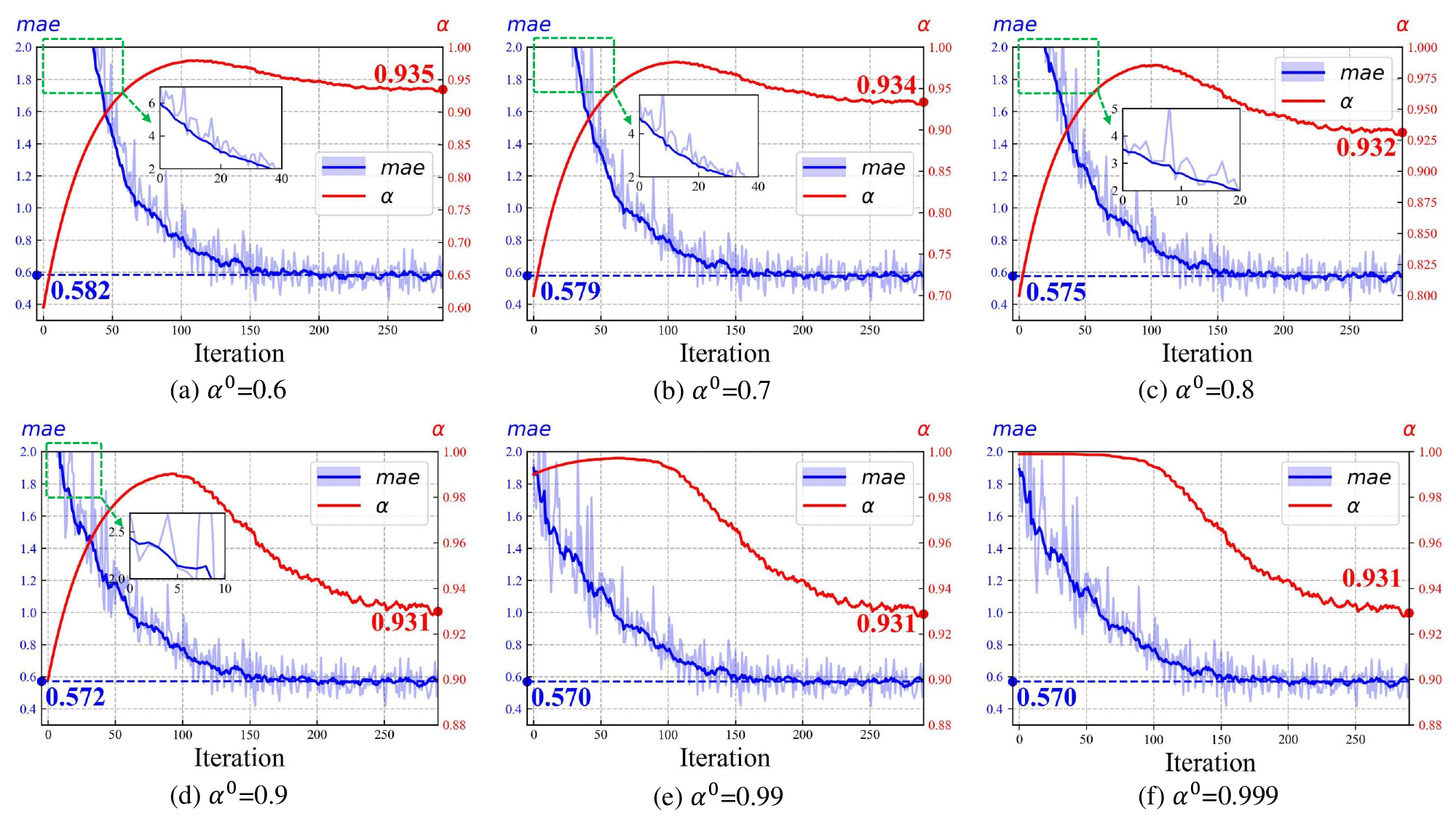}
    \caption{The curves of $mae$ and $\alpha$, across different initial value of $\alpha^0$. Results are derived from the second session of CUB. }
     \label{fig:abcdef}
\end{figure*}
\begin{figure}[t]
\centering
    \hspace{-0.4cm}\includegraphics[width=0.5\textwidth]{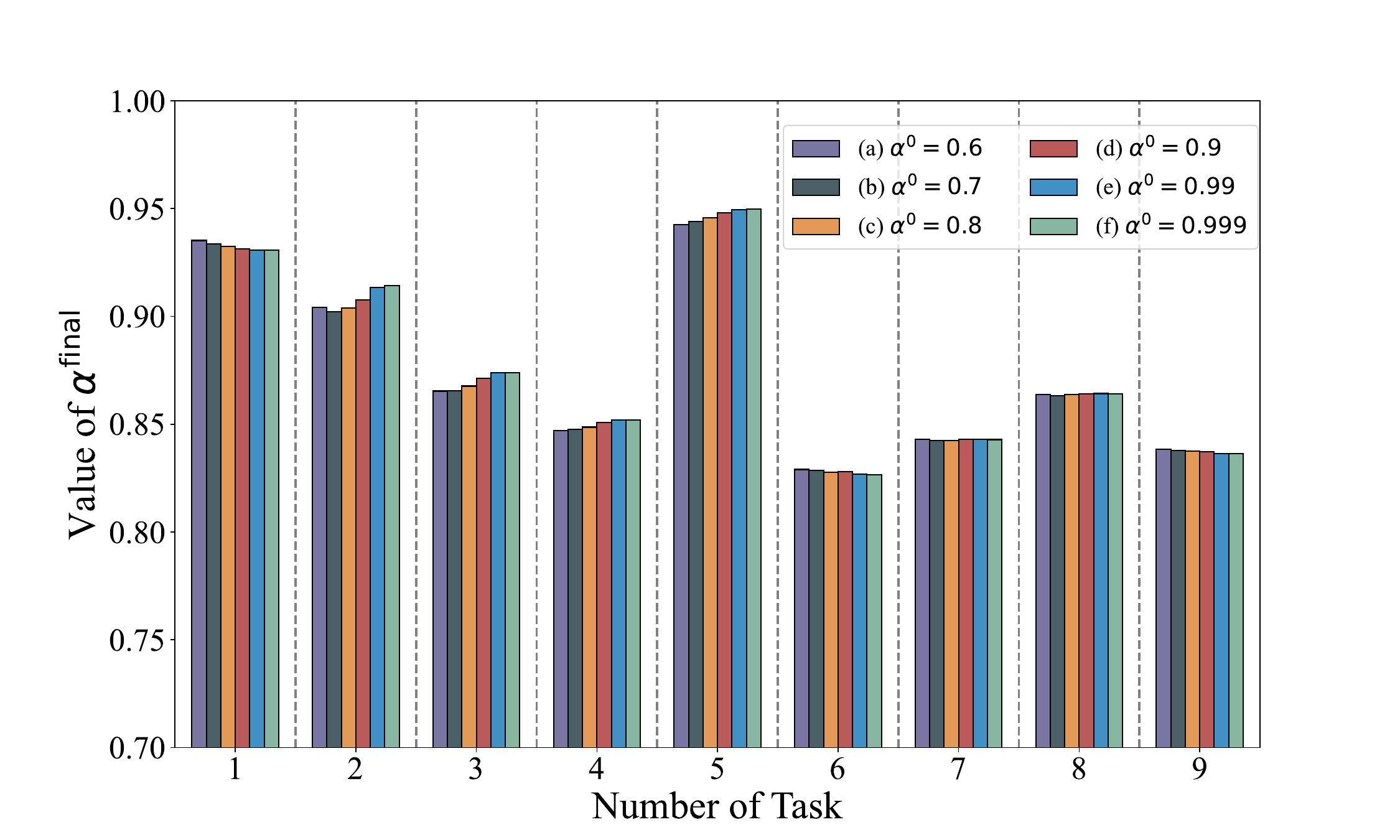}
    \caption{The value of $\alpha^{\mathrm{final}}$ across different settings. }
     \label{fig:barchart}
\end{figure}

We visualize the NKA update process across six settings as reported in Table~\ref{tab:nka}. Fig.~\ref{fig:abcdef} shows the update curves of $mae$ and $\alpha$ under various initial conditions. Both $mae$ and $\alpha$ consistently converge to a fixed value, suggesting this value represents the inherent correlation coefficient between tasks. As depicted in Fig.~\ref{fig:barchart}, although this coefficient varies across tasks, the model reliably converges to it regardless of initial conditions. This demonstrates that the NKA mechanism effectively reveals the inherent correlation in the data stream, promoting stable knowledge accumulation across different tasks.
This observation explains the significant performance improvement of PEARL, which is likely due to the NKA mechanism’s ability to address the distribution divergence between the downstream and pre-trained datasets by uncovering potential correlations.
% Consequently, PEARL may have potential advantages in cross-domain incremental learning.

The $mae$ metric indicates that a higher value reflects poorer knowledge preservation ability by the model.
As shown in the inset figures of Fig.~\ref{fig:abcdef}, the initial value of $mae$ is negatively correlated with the choice of $\alpha^0$. This suggests that the weighted mixing method for prompts effectively regulates knowledge mixing, thereby ensuring the NKA mechanism’s effectiveness.
The update curve of $\alpha$ exhibits an overshoot, consistent with a typical negative feedback response. This confirms the successful application of negative feedback regulation in our method. In the future, we will conduct further research on applying negative feedback regulation in deep learning.

%% file: tables/table1.tex
\begin{table*}[t]
  \centering
    \fontsize{9}{12}\selectfont
    \setlength{\tabcolsep}{1.1mm}
    \begin{tabular}{lrlrlrlrlrlrlrlrl}
    \toprule
    \multicolumn{1}{l}{\multirow{2}[2]{*}{Method}} & \multicolumn{2}{l}{CIFAR~20-tasks} & \multicolumn{2}{c}{CUB~20-tasks} & \multicolumn{2}{c}{IN-R~40-tasks} & \multicolumn{2}{c}{IN-A~10-tasks} & \multicolumn{2}{c}{Omni~10-tasks} & \multicolumn{2}{c}{VTAB~5-tasks} & \multicolumn{2}{c}{Average} \\
          & \multicolumn{1}{c}{ $\Bar{\mathcal{A}}$} & \multicolumn{1}{c}{$\mathcal{A}^{\mathcal{N}}$} & \multicolumn{1}{c}{ $\Bar{\mathcal{A}}$} & \multicolumn{1}{c}{$\mathcal{A}^{\mathcal{N}}$} & \multicolumn{1}{c}{ $\Bar{\mathcal{A}}$} & \multicolumn{1}{c}{$\mathcal{A}^{\mathcal{N}}$} & \multicolumn{1}{c}{ $\Bar{\mathcal{A}}$} & \multicolumn{1}{c}{$\mathcal{A}^{\mathcal{N}}$} & \multicolumn{1}{c}{ $\Bar{\mathcal{A}}$} & \multicolumn{1}{c}{$\mathcal{A}^{\mathcal{N}}$} & \multicolumn{1}{c}{ $\Bar{\mathcal{A}}$} & \multicolumn{1}{c}{$\mathcal{A}^{\mathcal{N}}$} & \multicolumn{1}{c}{ $\Bar{\mathcal{A}}$} & \multicolumn{1}{c}{$\mathcal{A}^{\mathcal{N}}$}  \\
    \midrule
    L2P   & 85.94  & 79.93  & 67.05  & 56.25  & 66.53  & 59.22  & 49.39  & 41.71  & 73.36  & 64.49  & 77.11  & 77.10  & 69.90  & 63.12  \\
    DualPrompt & 87.87  & 81.15  & 77.47  & 66.54  & 63.31  & 55.22  & 53.71  & 41.67  & 73.92  & 65.52  & 83.36  & 81.23  & 73.27  & 65.22  \\
    CODA-Prompt & 89.11  & 81.96  & 84.00  & 73.37  & 64.42  & 55.08  & 53.54  & 42.73  & 77.03  & 68.09  & 83.90  & 83.02  & 75.33  & 67.38  \\
    SimpleCIL & 87.57  & 81.26  & 92.20  & 86.73  & 62.58  & 54.55  & 59.77  & 48.91  & 79.34  & 73.15  & 85.99  & 84.38  & 77.91  & 71.50  \\
    ADAM + VPT-D & 88.46  & 82.17  & 91.02  & 84.99  & 68.79  & 60.48  & 58.48  & 48.52  & 81.05  & 74.47  & 86.59  & 83.06  & 79.07  & 72.28  \\
    ADAM + SSF & 87.78  & 81.98  & 91.72  & 86.13  & 68.94  & 60.60  & 61.30  & 50.03  & 80.53  & 74.00  & 85.66  & 81.92  & 79.32  & 72.44  \\
    ADAM + Adapter & 90.65  & 85.15  & 92.21  & 86.73  & 72.35  & 64.33  & 60.47  & 49.37  & 80.75  & 74.37  & 85.95  & 84.35  & 80.40  & 74.05  \\
    RanPAC & \underline{93.51}  & \underline{89.30}  & \underline{93.13}  & 89.40  & 75.74  & 68.75  & 64.16  & 52.86  & \underline{85.95}  & \underline{79.55}  & 92.56  & 91.83  & \underline{84.18}  & 78.62  \\
    HiDe-Prompt & 91.22  & \textbf{89.92}  & 89.75  & \underline{89.46}  & 76.20  & \textbf{74.56}  & 61.41  & 49.27  & 76.60  & 77.01  & 91.24  & 92.78  & 81.07  & \underline{78.83}  \\
    ESN   & 87.15  & 80.37  & 65.69  & 63.10  & 60.69  & 55.13  & 44.06  & 31.07  & 75.32  & 66.57  & 81.52  & 62.15  & 69.07  & 59.73  \\
    EASE  & 91.51  & 85.80  & 92.23  & 86.81  & \underline{78.31}  & 70.58  & \underline{65.34}  & \underline{55.04}  & 81.66  & 74.85  & \underline{93.61}  & \textbf{93.55}  & 83.78  & 77.77  \\
    \midrule
    PEARL (Ours) & \textbf{93.64}  &  {89.02} & \textbf{94.48}  & \textbf{89.65}  &  \textbf{79.54}     &  \underline{72.33}     & \textbf{67.41}  & \textbf{57.87}  & \textbf{86.87}
 &  \textbf{79.68}     & \textbf{96.52}  & \underline{93.02}  & \textbf{86.41} & \textbf{80.26}  \\
    \bottomrule
    \end{tabular}%
  \label{tab:addlabel}%
\caption{Comparison results on six benchmarks with \textbf{ViT-B/16-IN21K} as the backbone. Experiments are labeled as ``Dataset-$\mathcal{N}$-tasks'' where $\mathcal{N}$ represents the length of the data stream. ``IN-R'' is short for ImageNet-R, ``IN-A'' is short for ImageNet-A, and ``Omni'' is short for Omnibenchmark. Bold texts: the best results, underline texts: the second-best results.}
% The results come from ~\cite{zhou2024continual,EASE}. 
\label{tab:1}
\end{table*}%

%% file: tables/table5.tex
\begin{table}[htbp]
\setlength{\tabcolsep}{1mm}
  \centering
    \begin{tabular}{lccrlrl}
    \toprule
    \multicolumn{1}{l}{\multirow{2}[2]{*}{Method}} & \multicolumn{1}{c}{\multirow{2}[2]{*}{PE}} & \multicolumn{1}{c}{\multirow{2}[2]{*}{Mom}} & \multicolumn{2}{l}{IN-A 10-tasks} & \multicolumn{2}{l}{VTAB 5-tasks} \\
          &       &       & \multicolumn{1}{c}{ $\Bar{\mathcal{A}}$} & \multicolumn{1}{c}{$\mathcal{A}^{\mathcal{N}}$} & \multicolumn{1}{c}{ $\Bar{\mathcal{A}}$} & \multicolumn{1}{c}{$\mathcal{A}^{\mathcal{N}}$} \\
    \midrule
    \multirow{3}[1]{*}{L2P} &   --    &   --    & 53.36  & 43.45  & 80.84  & 61.40  \\
          &  \CheckmarkBold     &   --    &   \,\,\,3.00      &   \,\,\,1.18    &    \,\,\,5.36   &  \,\,\,2.76\\
          &  \CheckmarkBold     &  \CheckmarkBold     & \textbf{58.16}  & \textbf{49.18}  & \textbf{88.51}  & \textbf{69.74}  \\
    \midrule
    \multirow{3}[1]{*}{DualPrompt} &  --   &   --    & 57.05  & 46.61  & 83.03  & 66.32  \\
          &  \CheckmarkBold     &   --    &   56.71    &  45.69     &  80.22     &  63.51\\
          &  \CheckmarkBold     &  \CheckmarkBold     & \textbf{60.50}  & \textbf{50.69}  & \textbf{87.83}  & \textbf{77.27}  \\
    \midrule
    \multirow{3}[1]{*}{CODA-Prompt} &    --   &   --    & 59.67  & 47.33  & 81.79  & 84.75  \\
          &  \CheckmarkBold     &    --   &    15.93   &  \,\,\,3.55     &    42.46   & 32.05 \\
          &   \CheckmarkBold    &  \CheckmarkBold     & \textbf{61.83}  & \textbf{51.68}  & \textbf{84.82}  & \textbf{86.60}  \\
    \bottomrule
    \end{tabular}%

  \caption{Ablation study on SPA with \textbf{ViT-B/16-IN1K} as the backbone. ``PE'' is short for prompt encoder and ``Mom'' is short for momentum update with weight equals 0.9.}
  \label{tab:prompt_ablation}%
\end{table}%

%% file: tables/table4.tex
\begin{table}[t]
  \centering
  \setlength{\tabcolsep}{1.1mm}
    \begin{tabular}{lrlrl}
    \toprule
    \multirow{2}[0]{*}{\parbox[l]{2cm}{ Positional\\Encoding}} & \multicolumn{2}{c}{IN-A~10-tasks} &        \multicolumn{2}{c}{VTAB~5-tasks} \\
          & \multicolumn{1}{c}{ $\Bar{\mathcal{A}}$} & \multicolumn{1}{c}{$\mathcal{A}^{\mathcal{N}}$} & \multicolumn{1}{c}{ $\Bar{\mathcal{A}}$} & \multicolumn{1}{c}{$\mathcal{A}^{\mathcal{N}}$}  \\
    \midrule
    Continual       &  66.91     &   56.35    &  96.47     &   93.00     \\
    Segmented      &  \textbf{67.65}     &   \textbf{57.14}      &  \textbf{96.59}     & \textbf{93.07} \\
    \bottomrule
    \end{tabular}%
  \caption{Ablation study on the segmented positional encoding with \textbf{ViT-B/16-IN1K} as the backbone.}
  \label{tab:posi}%
\end{table}%

%% file: tables/table2.tex
\begin{table}[htbp]
  \centering
  \setlength{\tabcolsep}{1.5mm}
    \begin{tabular}{lrlrl}
    \toprule
    \multirow{2}[2]{*}{$\alpha^0$} & \multicolumn{2}{c}{Fixed $\alpha$} & \multicolumn{2}{c}{NKA $\alpha$} \\
          & \multicolumn{1}{c}{ $\Bar{\mathcal{A}}$} & \multicolumn{1}{c}{$\mathcal{A}^{\mathcal{N}}$} & \multicolumn{1}{c}{ $\Bar{\mathcal{A}}$} & \multicolumn{1}{c}{$\mathcal{A}^{\mathcal{N}}$}  \\
    \midrule
    0.60  & 60.04  & 44.83  & \textbf{81.56 } & \textbf{73.88 } \\
    0.70  & 61.41  & 47.96  & \textbf{84.47 } & \textbf{76.89 } \\
    0.80  & 68.79  & 63.74  & \textbf{87.28 } & \textbf{79.90 } \\
    0.90  & 86.32  & 82.15  & \textbf{89.73 } & \textbf{83.40 } \\
    0.99  & 82.21  & 70.31  & \textbf{91.26 } & \textbf{84.73 } \\
    0.999 & 80.64  & 66.16  & \textbf{91.33 } & \textbf{84.86 } \\
    \bottomrule
    \end{tabular}%
  % \label{tab:nka}%
  \caption{Ablation study on the NKA mechanism with \textbf{ViT-B/16-IN1K} as the backbone. The results are obtained on \textbf{CUB-200}, and $\mathcal{N}$ equals 10.}
  \label{tab:nka}
\end{table}%

%% file: m_Conclusion.tex
\section{Conclusion} \label{sec:conclusion}

This paper presents PEARL, an input-agnostic prompt method designed to address the issue of knowledge interference caused by the ``query-select'' mechanism of existing input-dependent prompt methods. Our method systematically manages and integrates task-specific knowledge through a global prompt, which helps mitigate catastrophic forgetting across incremental learning. Additionally, the proposed negative feedback based momentum update mechanism reveals potential correlations within the dataset, facilitating smooth and efficient knowledge accumulation. We hope our work can offer good insights into the field of CIL and provide some inspiration to other researchers. 

%%%%%%%%%%%%%%%%%%%%%%%%%%%%%%%%%%%%%%%%%%%%%%%%%%%%%%%%%%%%%%%%%%%%%%%%%%%%%%%%%%%%%%%%%
\section{Acknowledgments} \label{sec:acknowledge}
This work was supported by the National Natural Science Foundation of China (No. 62476056, 62076062, and 62306070) and the Social Development Science and Technology Project of Jiangsu Province (No. BE2022811). Furthermore, the work was also supported by the Big Data Computing Center of Southeast University.
%%%%%%%%%%%%%%%%%%%%%%%%%%%%%%%%%%%%%%%%%%%%%%%%%%%%%%%%%%%%%%%%%%%%%%%%%%%%%%%%%%%%%%%%%